\documentclass[fleqn,10pt]{wlscirep}
\usepackage[utf8]{inputenc}
\usepackage[T1]{fontenc}
\usepackage{listings}
\usepackage{xspace}
\definecolor{json-key}{rgb}{0.13,0.55,0.13}
\definecolor{json-value}{rgb}{0.25,0.49,0.94}
\definecolor{json-string}{rgb}{0.9,0.5,0.13}
\lstdefinelanguage{json}{
    basicstyle=\normalfont\ttfamily,
    numbers=left,
    numberstyle=\scriptsize,
    stepnumber=0,
    numbersep=8pt,
    showstringspaces=false,
    breaklines=true,
    frame=lines,
    backgroundcolor=\color{white},
    stringstyle=\color{json-string},
    commentstyle=\color{gray},
    keywordstyle=\color{json-key},
    morekeywords={true,false,null},
    keywords=[2]{},
    keywordstyle=[2]\color{json-value},
    literate=
     *{0}{{{\color{json-value}0}}}{1}
      {1}{{{\color{json-value}1}}}{1}
      {2}{{{\color{json-value}2}}}{1}
      {3}{{{\color{json-value}3}}}{1}
      {4}{{{\color{json-value}4}}}{1}
      {5}{{{\color{json-value}5}}}{1}
      {6}{{{\color{json-value}6}}}{1}
      {7}{{{\color{json-value}7}}}{1}
      {8}{{{\color{json-value}8}}}{1}
      {9}{{{\color{json-value}9}}}{1}
}
\newcommand{\ourmethod}{\textit{Open Datasheets}\xspace}

\title{Open Datasheets: Machine-readable Documentation for Open Datasets and Responsible AI Assessments}

\author[1,*]{Anthony Cintron Roman}
\author[2]{Jennifer Wortman Vaughan}
\author[3]{Valerie See}
\author[4]{Steph Ballard}
\author[5,+]{Kevin Xu}
\author[1,+]{Jehu Torres}
\author[1,+]{Caleb Robinson}
\author[1]{Juan M. Lavista Ferres}
\affil[1]{Microsoft, AI For Good Research Lab, USA}
\affil[2]{Microsoft, Research, USA}
\affil[3]{Microsoft, Office of Responsible AI, USA}
\affil[4]{Microsoft, Open Innovation, USA}
\affil[5]{GitHub, USA}
\affil[*]{anthony.cintron@microsoft.com}
\affil[+]{these authors contributed equally to this work}

\begin{abstract}
This paper introduces a no-code, machine-readable documentation framework for open datasets, with a focus on responsible AI (RAI) considerations. The framework aims to improve comprehensibility, and usability of open datasets, facilitating easier discovery and use, better understanding of content and context, and evaluation of dataset quality and accuracy. The proposed framework is designed to streamline the evaluation of datasets, helping researchers, data scientists, and other open data users quickly identify datasets that meet their needs and organizational policies or regulations. The paper also discusses the implementation of the framework and provides recommendations to maximize its potential. The framework is expected to enhance the quality and reliability of data used in research and decision-making, fostering the development of more responsible and trustworthy AI systems.
\end{abstract}
\begin{document}

\flushbottom
\maketitle
\thispagestyle{empty}

\section*{Introduction}

Machine-readable documentation, which refers to documentation that is structured in a format easily readable and processed by computers, plays a vital role in improving the accessibility, comprehensibility, and usability of open datasets. It has the potential to enable easier discovery and use of datasets, facilitate a better understanding of the content and context of a dataset, simplify the integration of datasets from various sources, and help with the evaluation of the quality and accuracy of datasets\cite{heger2022understanding}. With the exponential growth of open data, machine-readable metadata is becoming increasingly important for making this data easily discoverable and useable\cite{contaxis2022ten}.

As open data is progressively employed in AI applications\cite{mckinsey_opendata,BRINKHAUS2023102542}, it is imperative for researchers to consider the potential implications of these open datasets from a responsible AI (RAI) perspective\cite{heger2022understanding}. It is widely recognized that AI systems can inherit biases and other flaws from the data they are trained on \cite{heger2022understanding,buolamwini2018gender}, and with AI continuing to permeate various aspects of our daily lives, the transparency, reliability, trustworthiness, and fairness of AI systems becomes paramount. This begins with the quality of the data they are trained on\cite{paullada2021data}. Understanding the data being used in an AI system is crucial to uphold responsible AI implementations, including its source, the methodologies used in its collection, and other relevant factors discussed in this paper.

Evaluating a dataset involves multiple distinct goals or aspects. One is to determine if a dataset complies with organizational policies and/or regulations, such as ensuring appropriate consent was obtained. The other is to ascertain if a dataset is suitable for the task at hand, such as whether the labels are valid proxies for the real-world phenomena a  model is being designed to predict. While researchers can often evaluate an open dataset and make determinations about some of its characteristics, such as demographic distribution, as part of their exploratory data analysis (EDA), there are properties that cannot be learned through data exploration, such as exactly how demographic information was obtained from data subjects or whether and how their consent was obtained. This makes searching for and evaluating datasets an overwhelming, if not impossible, task. It can lead to investing a significant amount of time in evaluating a dataset only to realize later that it was not collected properly and/or does not have a proper representation from an RAI perspective to meet their organizational policies, among other considerations. Such mistakes can be costly, with potential harm to individuals, and significant losses for organizations\cite{iapp_webscrape_data,price2019privacy,trivedi2019risks,forbes_cost_bad_data,entrepreneurs_cost_bad_data,hbs_data_integrity_importance,gartner_cost_bad_data}.

Consider a researcher working on a project related to speech recognition technology. They need a dataset of audio recordings to train their AI model. Without proper documentation, they start searching the web for open datasets that might be suitable for their project. They find a dataset that claims to contain a large collection of audio recordings. However, there is no detailed information about how the dataset was collected, who the speakers are, or any ethical considerations that were considered during the data collection process. The researcher is unsure if the dataset includes recordings of individuals without their consent or if it violates any privacy regulations. They also cannot easily determine whether the conditions under which the recordings were collected match the scenarios they are interested in modeling. 

To ensure responsible AI practices, the researcher needs to evaluate the dataset thoroughly. They spend a significant amount of time manually examining the dataset, trying to find any clues about its quality, ethical considerations, and compliance with regulations. However, due to the lack of proper documentation, they cannot make an informed decision. After investing a considerable amount of time, the researcher realizes that the dataset does not meet their organizational policies and cannot be used for their project. They have to start the search process all over again, wasting valuable time and resources.

Therefore, the ability to filter out datasets that do not meet organizational policies or the specific objectives of the task at hand, can create significant time and cost savings. As a first step, automating and/or simplifying the evaluation of datasets through metadata can help researchers quickly identify datasets that do not meet their needs and/or organizational policies. However, it's important to note that this does not eliminate the need for EDA on a dataset that will be used to train production models, i.e. identifying properties of the dataset that can be learned through data exploration, like demographics. Instead, it aims to reduce the amount of time invested in filtering datasets that do not meet broader criteria.

Furthermore, given the time-consuming and costly nature of finding the right dataset, an applied documentation framework becomes indispensable in making the process of discovering and evaluating open datasets more efficient and cost-effective. By implementing a comprehensive documentation framework, we can help address biases, enhance transparency, and promote responsible AI practices. Such a framework empowers researchers and practitioners to thoroughly understand the data they are using, enabling them to assess its suitability and identify any potential biases or limitations. Additionally, it streamlines the dataset discovery process by providing clear and standardized documentation, reducing the time and effort required to find and evaluate datasets.

By emphasizing the significance of a reliable documentation framework, we can elevate the quality and trustworthiness of AI systems, fostering responsible and ethical AI practices. In light of this, we propose a no-code, machine-readable documentation framework to assist in the evaluation of open datasets, improve usability, and consider responsible AI aspects. Our contributions include:
\begin{itemize}
    \item Publication of the proposed JSON-based metadata framework for open datasets on GitHub; \url{https://github.com/microsoft/opendatasheets-framework}
    \item Implementation of a public no-code solution hosted on GitHub Pages to generate and evaluate this metadata; \url{https://microsoft.github.io/opendatasheets}
    \item Discussion of recommendations to maximize the potential of this framework, improving efficiency, and providing transparency for consumers of open datasets
\end{itemize}

\section*{Related Work}

In recent years there have been calls for comprehensive documentation of the datasets used to train and evaluate AI systems\cite{gebru2021datasheets,heger2022understanding,holland2020dataset,bender2018data,hutchinson2021towards,raji2019ml}. Several data documentation frameworks and tools have been proposed with the goal of encouraging thoughtful reflection on datasets and transparency about their makeup and creation process.  Datasheets for datasets \cite{gebru2021datasheets} is a documentation framework designed to encourage dataset creators to reflect on the implicit and explicit choices behind their data and to enable those interested in using a dataset to train or evaluate their system to make more informed decisions about its appropriateness.  Developed concurrently, data nutrition labels \cite{holland2020dataset,chmielinski2022dataset} also include information about a dataset’s provenance and makeup.  The authors created a publicly available tool to create a nutrition label that highlights usage restrictions and potential harms that may arise from the dataset’s use.  Data statements for natural language processing \cite{bender2018data} are designed specifically for natural language datasets and include specialized information such as speaker demographics and language variety.  More recently, researchers at Google released the Data Cards Playbook\cite{pushkarna2022data},  while researchers at Microsoft released the Aether Data Documentation Framework, a variant of datasheets for datasets adapted to meet the needs of industry practitioners\cite{heger2022understanding}.

While research shows that well-designed data documentation can be effective at helping to identify ethical issues\cite{boyd2021datasheets}, good documentation can be challenging to create.  In a study of industry practitioners’ data documentation perceptions, needs, challenges, and desiderata, Heger et al. \cite{heger2022understanding} found that practitioners creating dataset documentation had trouble making connections between the questions they were asked to answer and their RAI implications and difficulty providing information that someone unfamiliar with their datasets would need to understand the data. Based on their findings, they derived seven design requirements that data documentation frameworks should satisfy. Briefly, these include 1) making the connection to RAI more explicit; 2) making data documentation frameworks practical; 3) adapting data documentation frameworks to different contexts; 4) supporting simple tasks with automation without automating away responsibility; 5) clarifying the target audience for the documentation; 6) standardizing and centralizing data documentation; and 7) integrating data documentation frameworks into existing tools and workflows.

Inspired by and building on this line of work, our proposed framework, Open Datasheets, introduces a machine-readable metadata format for open datasets. Drawing on questions included in datasheets for datasets \cite{gebru2021datasheets} and the Aether data documentation template \cite{aether_doc}, it includes detailed information about the datasets, including responsible AI considerations. Furthermore, it adheres to the seven design principles of Heger et al.\cite{heger2022understanding}, as discussed in the design and implementation sections. Unlike existing tools, the framework aims to integrate with existing open platforms and support conversion to other standard formats, such as JSON-LD, providing an applied and efficient open framework that is user-friendly for non-developers and other open data publishers and users. 

Other documentation frameworks and formats, such as those utilized by HuggingFace\cite{mcmillan2021reusable,lhoest2021datasets}, were also considered but not included here due to their specialization and association with proprietary platforms. The evaluation focused on flexible published frameworks that can be adapted, are open, and align closely with the 7 design principles proposed by Heger et al.\cite{heger2022understanding}.

\subsection*{Methods}

The approach for designing and implementing the \ourmethod framework involved a multistage process that incorporated feedback from various dataset producers. This approach also included researching existing data documentation frameworks \cite{gebru2021datasheets,holland2020dataset,chmielinski2022dataset,bender2018data,pushkarna2022data} and studying the challenges and opportunities identified in previous studies \cite{gebru2021datasheets,heger2022understanding,holland2020dataset,bender2018data,hutchinson2021towards,raji2019ml} on data documentation.

Our primary research aim was to identify the areas that needed improvement in practical implementation for open data documentation. To achieve this, we collaborated with data scientists and researchers from the Microsoft AI for Good Research Lab \cite{MicrosoftAIForGoodTeam}, who leverage and publish open datasets for their efforts to solve societal challenges with AI. Through hands-on experience and collaboration, we aimed to fill the gaps in existing frameworks and documentation practices.

The second aim of our approach was to leverage an existing documentation standard to facilitate adoption, avoid reinvention, and focus on bridging the gaps identified in previous studies on responsible AI documentation. We selected a format and standard based on criteria such as simplicity, user usability, and practicality to support non-developers, researchers, and developers.

The third aim was to provide a framework that includes no-code tooling, resources, and guidance that facilitate the understanding of the importance and application of documentation for open datasets.

We engaged with a diverse cohort of open dataset producers from Microsoft, including data scientists, researchers, analysts, and program managers involved in dataset publication and documentation. Their varying perspectives provided valuable insights.

\subsubsection*{Initial Iteration}

The initial evaluation phase occurred during the early development of the \ourmethod framework web application. We employed the following methodologies:
\begin{itemize}
	\item Case Study Analysis: We applied the \ourmethod framework to three distinct open data publications hosted on GitHub \cite{clinical_sum_github,clandestino_github,rtp_lx_github}. These projects were selected to represent a range of producers, including non-developer researchers and data scientists.
	\item Qualitative Feedback: We collected insights from publishers through observations and conversations. Their feedback helped us understand the importance of guidance for responsible AI documentation and determine what information and features should be included.
	\item Usability Testing: We conducted usability testing to identify challenges in the absence of automated features and inline guidance.
\end{itemize}
\subsubsection*{Challenges}

Identified Publishers experienced difficulties in several areas:
\begin{itemize}
	\item Comprehending the value of responsible AI documentation.
	\item Deciding on the required information for adequate documentation.
	\item Navigating the framework without inline guidance or automation tools.
\end{itemize}
 
These observations were consistent with findings from Heger et al.’s research \cite{heger2022understanding}, which highlighted similar challenges in documentation tools.
\subsubsection*{Iterative Improvement}

In response to the feedback from the initial iteration, we refined the framework to provide a better balance of features:
\begin{itemize}
    \item Enhanced Inline Guidance: We improved the inline help components to instruct users on creating effective responsible AI documentation, providing clear examples and definitions.
	\item Automation Features: We introduced automation tools to streamline the documentation process. These tools can infer and prefill certain parts of the datasheets based on the data itself, reducing manual input.
	\item User Documentation: We provided comprehensive user documentation, including a step-by-step guide and resources, to help dataset producers navigate the framework and understand best practices for responsible AI documentation.
\end{itemize}
\subsubsection*{Design and Implementation Insights}

The refined design and implementation were tailored to address the key challenges identified, resulting in a more intuitive and resourceful framework. Automation and guidance components were fine-tuned to ensure dataset producers receive support throughout the documentation process.

The methods employed serve to refine the \ourmethod framework into a tool that accommodates varied user needs while emphasizing the importance of robust documentation for open datasets. Each step of the process has been scrutinized and enhanced to align with feedback and prior research outcomes.

\section*{Design}
The \ourmethod framework is designed to streamline the documentation of open datasets. It aims to foster the inclusion of crucial information that assists users in comprehending potential biases, privacy concerns, and other elements of responsible AI. This is achieved by striking a balance between automation and thoughtful deliberation in the documentation process \cite{heger2022understanding}. The ultimate objective is to enhance data documentation for open datasets, improve their discoverability, and promote their reliability, fairness, and transparency when used for AI models.

The \ourmethod format specification is rooted in the Datapackage standard, a product of the Frictionless Data project\cite{FrictionlessData2023}. This open-source initiative offers tools and standards for data management. The Datapackage specification, accessible on GitHub\cite{frictionlessdata_spec_github}, is a user-friendly JSON based format that acts as a container for describing a set of data. Also, it supports different data types, including complex ones such as tabular and geographic data, allowing users to document the basic information of a dataset in a way that is tailored to their specific needs. This standard has been widely adopted on open data platforms such as \url{https://data.world}.

The \ourmethod framework utilizes the Datapackage format specifications to consistently document basic dataset details. This approach enables users to easily comprehend the information on the datasheets by avoiding specialized terminology and adhering to the standard language and usage of the JSON format. This format is machine-readable and seamlessly integrates into data workflows, thereby enhancing data discovery, access, and usage. The adoption of these specifications fosters compatibility and uniformity in the realm of data documentation.

\begin{lstlisting}[caption={Datapackage Sample},label=samplecode2,language=json]
{
  # general "metadata"
  "name" : "a-unique-human-readable-and-valid-url-identifier",
  "title" : "A descriptive title",
  "licenses" : [...],
  "sources" : [...],
  # list of the data resources
  "resources": [
    {
      ... resource information ...
    }
  ]
}
\end{lstlisting}

Although the Datapackage specification is beneficial for general data documentation, it may not suffice for organizations that need to comply with regulatory standards and ensure responsible AI practices. To cater to this need, the \ourmethod framework incorporates concepts from ``Datasheets for Datasets''\cite{gebru2021datasheets}. These concepts, such as detailed descriptions of a dataset's privacy implications to encourage careful reflection, are integrated into a machine-readable format. This additional information is crucial for organizations to make informed decisions about dataset usage and ensure compliance with their policies.

Moreover, the \ourmethod Framework adheres to the 7 design principles outlined in section 5.2 of Heger et al research\cite{heger2022understanding}. It explicitly connects data documentation with responsible AI through inline guidance based in part on Microsoft’s Aether data documentation template\cite{aether_doc}. The framework is practical, integrating with GitHub and providing a user-friendly interface. It adapts to different contexts, allowing customization. Automation is balanced with responsibility, automating foundational metadata extraction while guiding users on responsible AI metadata. It clarifies the target audience to data publishers, focusing on potential users of open datasets. Standardization is achieved through a machine-readable format, and integration with GitHub promotes collaboration. The framework seamlessly integrates into existing tools and workflows, associating documentation with the data publication and development lifecycle on GitHub.

Furthermore, with the emergence of Large Language Models (LLMs), machine-readable interpretation of complex documentation has greatly advanced. The \ourmethod framework enables organizations to leverage these advancements by documenting datasets with more descriptive information, allowing for more effective analysis and interpretation. This includes automated interpretation as a preliminary step to exclude datasets that do not align with organizational policies.

\begin{lstlisting}[caption={RAI documentation Sample},label=raidoc,language=json]
"privacy": [{
     "sensitivity": {
         "description": "sensitivity types description",
         "types": [
             {
                 "name": "political opinions",
                 "description": "description of the content related to this type"
             }
         ]
     },
     "confidentiality": {
         "path": "https://microsoft.github.io/opendatasheets/confidentiality",
         "description": "description of the process to ensure the confidentiality of the data subjects"
     }
 }],
"procedures": {
    "collection": [{
            "description": "Procedure description",
            "path": "",
            "contributors": [],
            "methods": [
                {
                    "name": "focus group",
                    "description": "focus group description",
                    "path": "/focusgroup.txt"
                }
            ],
            "consent": [
                {
                    "title": "consent form",
                    "description": "consent form description",
                    "path": "/consentform.pdf"
                }
            ]}
    ]}
\end{lstlisting}
\section*{Implementation}

\begin{figure}[ht]
  \centering
  \includegraphics[width=0.5\textwidth,height=0.3\textheight]{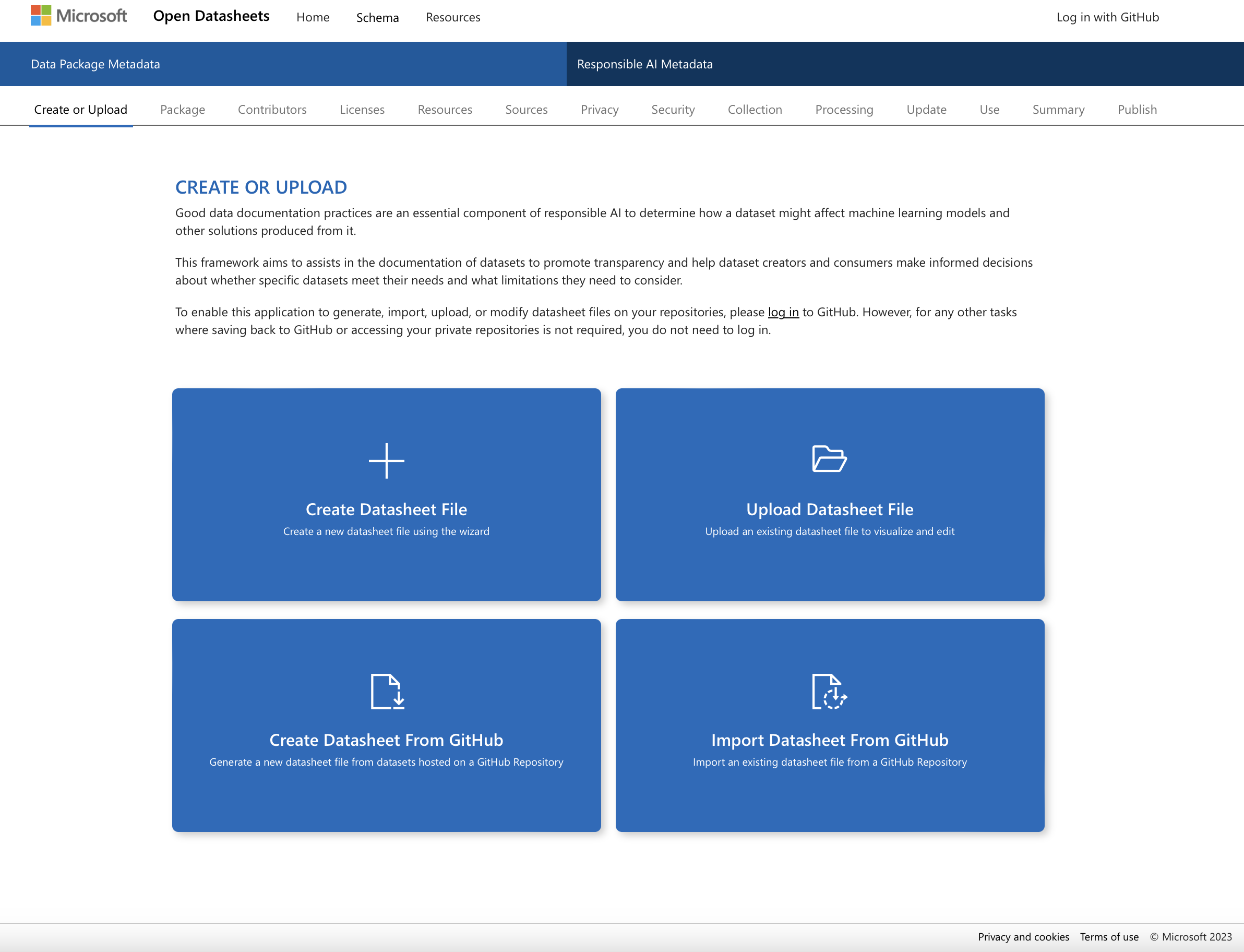}
  \caption{Open Datasheets Web Application (https://microsoft.github.io/opendatasheets)}
  \label{fig:Open_Datasheets_Web_App}
\end{figure}

To achieve a no-code solution, the framework implements a user-friendly web application on GitHub Pages (Figure \ref{fig:Open_Datasheets_Web_App}), a managed service by the GitHub Platform to host static websites on GitHub repositories. This implementation ensures the longevity and support of the framework. The web application features a wizard-style interface that assists in standardizing the documentation according to the framework's metadata format. It includes metadata parsers for common data file formats on the GitHub platform\cite{roman2023open}, such as CSV, TSV, JSON, and others. These parsers extract metadata related to the file structure, including field names, types, and sample values. General metadata about the data file, such as filename, encoding, and size, is also extracted. This approach allows data publishers to focus solely on filling in the responsible AI metadata.

The integration with the GitHub Platform, one of the largest open data platforms\cite{roman2023open}, promotes openness and community. Currently, the web application enables seamless documentation of datasets on GitHub, enhancing discoverability and transparency for datasets hosted on the platform. However, the \ourmethod framework is not limited to documenting datasets exclusively on GitHub. It provides the flexibility to create documentation for datasets and download the metadata file for publishing on any preferred platform. Moreover, the framework is openly available on GitHub, ensuring easy accessibility for data publishers.

The \ourmethod framework simplifies the data documentation process for data publishers and promotes the inclusion of responsible AI elements. By automating the extraction of foundational metadata from data files and providing visual guidance for responsible AI considerations, the framework reduces the time and effort required for extensive documentation. This addresses the reluctance of data publishers to write lengthy documentation, as highlighted in Heger et al research \cite{heger2022understanding}. 

For data users, the \ourmethod framework offers detailed and machine-readable documentation, enabling informed decisions regarding dataset selection and use. The framework encourages data publishers to document potential biases and privacy implications associated with the dataset. This documentation empowers data users to evaluate the reliability and fairness of the dataset. The comprehensive documentation encouraged by the framework enhances transparency and trustworthiness, allowing data users to assess the dataset's quality, understand its limitations, and ensure it aligns with their ethical standards and requirements. 

The standardized, machine-readable documentation also saves time and effort for data users, enabling them to programmatically evaluate the dataset's suitability for their needs and seamlessly integrate it into their organizational workflow for automated initial evaluation of usability and potential issues. Data users can also assess the documentation using the web application, which provides a user-friendly graphical representation of the metadata. 

The \ourmethod framework metadata format comprises two main sections: the dataset foundational metadata and the responsible AI metadata. The dataset foundational metadata extends the Datapackage standard and includes essential information such as the dataset's name, title, licenses, sources, and a comprehensive list of the data resources within the package. On the other hand, the responsible AI metadata builds on the concepts from "Datasheets for datasets" \cite{gebru2021datasheets} and Microsoft's Aether Data Documentation Template \cite{aether_doc}. It provides information about the data's origin, processing methods, privacy implications, potential biases, and other relevant aspects. These components are further discussed in the following sections, specifically the \textit{Datapackage} and \textit{Responsible AI} sections.

\subsection*{Datapackage}

The Datapackage module describes the foundational metadata of the framework. This module includes the package description, licensing information, contributors, resources, and sources associated with the dataset. This helps in understanding the composition of a dataset and the structure of the data assets contained within the datasets, which is crucial. The datapackage provides essential information for this purpose.

\begin{description}
\item[Package]includes descriptors like name and description, version and creation date.
\item[Licensing]provides the terms under which the data is shared, an important factor in determining appropriate use cases for the given dataset.
\item[Contributors]provides information about the people or organizations that created and/or collected, and are sharing the dataset, which assists in identifying the provenance of the dataset.
\item[Resources]describe the list of data files included in the datapackage, including their data fields and value types in the case of tabular data.
\item[Sources]describe the list of data sources utilized for the creation and/or collection of the dataset, which also assist in identifying the provenance of the dataset.
\end{description}

\subsection*{Responsible AI}

The Responsible AI module of the \ourmethod framework focuses on documenting aspects that can impact user privacy, security, and biases that can undermine the trustworthiness of AI systems when used for training AI models. Evaluating datasets is crucial for creating responsible AI systems because the quality of the data directly affects the accuracy and fairness of the resulting system. Biases in the data can lead to biased outcomes, which can have serious consequences for individuals and society\cite{obermeyer2019dissecting,chen2023human,mckinsey_bias}. For instance, if an AI system used for hiring decisions is trained on biased data, it may perpetuate discrimination against certain groups.

To address this issue, organizations must evaluate their datasets for potential biases and take steps to mitigate them. The \ourmethod framework provides a standardized approach to document and share dataset information, including metadata that aids in the evaluation of responsible AI. This metadata includes details about the data collection process, preprocessing or cleaning steps, and any identified biases.

By incorporating this metadata into their responsible AI workflows, organizations can automate much of the initial evaluation process and ensure that open datasets align with their responsible AI policies. This not only saves time and resources but also contributes to promoting fairness, reducing bias, and increasing transparency in the resulting AI system for all users.

However, it is important to note that human review will still be necessary to make decisions based on organizational policies. The \ourmethod framework is not a substitute for human judgment but rather a tool to assist in the evaluation of responsible AI. By combining human expertise with machine-readable metadata, organizations can create more responsible and trustworthy AI systems.

With this objective in mind, the framework encompasses the following responsible AI areas:

\begin{itemize}
    \item Privacy
    \item Data Access
    \item Collection Procedures
    \item Processing Procedures
    \item Update Procedures
    \item Use Cases
\end{itemize}

\subsubsection*{Privacy and Use Terms}

To strike a balance between the advantages of open data and safeguarding individuals' privacy rights while promoting ethical data practices, it is crucial to evaluate the privacy implications of open datasets. The \ourmethod framework promotes thoughtful consideration of a dataset's privacy implications by integrating metadata on confidentiality, data sensitivity, and usage terms for the data.

\begin{description}
\item[Confidentiality]is a key aspect of privacy assessments. It refers to the protection of personal information from unauthorized access, use, or disclosure\cite{nist_confidentiality_definition}. Therefore, having this incorporated into the framework helps the users of the dataset understand the measures that were taken, if any, by the publishers to avoid the disclosure of personal information in their dataset.

\item[Data Sensitivity]in the context of privacy, refers to the level of sensitivity or potential harm associated with certain types of personal information included in a dataset. This includes sensitive attributes such as race, sex, religion, sexual orientation, health information, financial data, and other personally identifiable information (PII)\cite{iapp_cpra_sensitive_categories,ico_gdpr_sensitive_categories}. These sensitive attributes have the potential to cause harm or discrimination if mishandled, accessed, or disclosed without proper consent or safeguards.

\item[Use Terms]define the terms and conditions under which the data can be accessed, used, and shared. These terms help ensure that the data is used in a responsible and ethical manner, and that the privacy and confidentiality of individuals' personal information is protected. Use Terms can include restrictions on the types of analyses that can be performed on the data, limitations on the sharing or redistribution of the data, and requirements for obtaining consent or anonymizing the data before use. By defining these terms, data providers can help prevent the misuse or unauthorized access of sensitive data and ensure that the data is used for its intended purpose.

\end{description}

It is essential to evaluate a dataset for privacy implications because not only does it help protect the privacy of individuals whose data is being used, but also because many countries have laws and regulations that mandate organizations to safeguard individuals' privacy. By assessing a dataset for privacy implications, organizations can ensure that they comply with these laws and regulations.

\subsubsection*{Data Access}

Including information on how to access a dataset is also essential for privacy and security considerations. One key reason is to clearly communicate whether the data can be accessed anonymously or if registration is required. This information helps potential users understand how to access the data and determine if they meet the necessary criteria to access it.

Furthermore, explaining the intended use of the dataset is an important aspect of data access documentation. By providing information on how the data can be accessed, dataset documentation allows users to assess whether the dataset aligns with their specific needs and purposes. For example, if the dataset is intended for academic research, requiring registration may be necessary to ensure that it is accessed only by qualified individuals or institutions.

Another important reason for specifying data access methods in the dataset documentation is to establish clear terms of use. This helps prevent non-qualified access or misuse of the data, preserving its integrity and privacy.

In line with these considerations, the \ourmethod framework incorporates a module that enables data publishers to clearly specify how people can access the data. Additionally, it enables dataset providers to establish terms of use for the data and maintain data security. By providing this information, data providers can promote transparency, accountability, and responsible data usage.

\subsubsection*{Collection Procedures}
Building upon the importance of evaluating datasets for privacy implications, the \ourmethod framework also recognizes the significance of documenting data collection procedures as a crucial aspect of responsible AI considerations. By documenting these procedures, data users can gain insight into how and what data was collected. This promotes transparency and builds trust when using the data to train an AI system, reducing the potential for bias or discrimination in the system. 

Additionally, documenting data collection procedures enables the replication of the process in the future, which is vital for researchers to validate the results of an AI system trained on that data. By replicating the study and following the documented collection procedures, researchers can confirm that the system is functioning as intended. If the replicated study yields different results, it suggests the need for further investigation to identify potential issues or biases within the original dataset or data collection procedures.

Furthermore, documenting data collection helps data users ensure that the process adheres to ethical and legal standards. The framework takes into consideration the methods, consent forms utilized for data subjects and the contributors of the collection procedures.

\begin{description}
    \item[Methods] describe the approach and instruments used to collect the data such as surveys, interviews, websites and others.
    \item[Consent] is necessary to obtain when collecting personal data, especially sensitive data. Obtaining consent ensures compliance with privacy laws, respects individuals' rights, and promotes transparency and trust in data collection practices. For data users, understanding if consent was obtained provides insights into whether the data was collected ethically and in adherence to privacy laws and regulations.
    \item[Contributors] describe who performed the data collection, if an individual, an organization or a third-party. This helps data users understand if the data publisher was not the entity that collected the data.
\end{description}

\subsubsection*{Processing Procedures}
Documenting data pre-processing or processing procedures is important for reproducibility, transparency, accountability, and quality control, in addition to documenting collection procedures. This is because understanding how the data was processed makes it easier to identify errors or issues with the data. For example, let's say a dataset was pre-processed by removing all missing values without any imputation. If this pre-processing step is not documented, it may not be immediately apparent that some data points were removed, potentially leading to biased results or incorrect conclusions. Overall, documenting data preprocessing procedures is crucial for ensuring responsible and reliable AI systems. 

As part of the data processing documentation, the \ourmethod framework includes processing methods and contributors to this step.

\begin{description}
    \item[Methods] describe the approach and methodology used to process the data such as aggregation, anonymization, labeling and others.
    \item[Contributors] describe who performed the data processing, if an individual, an organization or a third-party. Like the data collection contributors, data users can understand who processed the data.
\end{description}

\subsubsection*{Update Procedures}

Datasets can be categorized as either static or periodically updated. A static dataset remains unchanged over time, while a periodically updated dataset undergoes regular updates to incorporate new data or revisions. When choosing between these options, researchers consider their specific objectives. If the research requires analyzing historical data or examining trends over a specific period, a static dataset may be suitable. Conversely, if the research involves analyzing real-time or evolving phenomena, a periodically updated dataset would be more appropriate. By documenting the procedures for updating the dataset, data users can gain insights into the timing and nature of the updates. This documentation promotes transparency and builds trust in the data, as users can verify its recency and reliability.

To facilitate this process, the \ourmethod framework incorporates descriptors to identify whether a dataset is static or updated, the periodicity of updates, the method and versioning procedures, and the contributors for the update procedures.
\begin{description}
\item[Is Updated] identifies whether a dataset is static or periodically updated.
\item[Periodicity procedure] describes the schedule or frequency of updates for the dataset.
\item[Method] describes the approach for updating the dataset, such as incrementally or a full refresh of the dataset.
\item[Versioning procedure] indicates how the data is versioned and how often.
\item[Contributors] describe who performs the data updates, whether it is an individual, an organization, or a third-party.Like the data collection and processing contributors, data users can understand who is responsible for updating the data.
\end{description}

\subsubsection*{Use Cases}

The framework also incorporates the concept of use cases into the documentation. This involves documenting how a dataset can be used and what it cannot be used for, which is crucial for responsible AI. By understanding the potential uses and limitations of a dataset, it becomes easier to comprehend its boundaries. This documentation plays a vital role in preventing unintended consequences that may arise from utilizing the data in unintended ways. Additionally, it helps identify ethical considerations that must be taken into account when using the data in an AI system.

Overall, documenting the permissible and impermissible uses of a dataset is a crucial step in ensuring responsible AI systems. It provides guidelines for utilizing the data in a responsible and ethical manner, while also highlighting clearly ill-advised uses.

As a result, the \ourmethod framework considers including use cases or examples of use for the dataset.

\section*{Conclusion}

In this publication, we have proposed and explored the integration of Responsible AI (RAI) documentation into a machine-readable metadata framework for open datasets. Our contribution includes a no-code web application designed to simplify the documentation process. The \ourmethod framework we have developed provides a robust foundation for dataset documentation, offering simplicity and flexibility that caters to data publishers working with diverse data types and requiring a customizable data management approach.

The framework covers core areas crucial for evaluating RAI considerations in a machine-readable format. These areas include privacy, collection and processing procedures, limitations of use, and other relevant aspects. We believe that this framework can be integrated into automated workflows to determine whether a dataset satisfies an organization's compliance criteria and warrants human review, or if it should be automatically discarded.

By incorporating the concepts of ``Datasheets for Datasets''\cite{gebru2021datasheets} and the design principles of Heger et al. \cite{heger2022understanding} into our framework, we have developed a more comprehensive and transparent approach to sharing and documenting data. This can lead to improved decision-making and the development of more trustworthy AI systems.

Looking ahead, we propose several areas for future work on the \ourmethod framework. These include extending metadata extraction automation to incorporate more data types, evaluating integration with data governance frameworks, fostering collaboration and community building around the data documentation framework, and automating the validation of the richness and quality of the free-form text for RAI evaluations.

In conclusion, the \ourmethod framework is a significant contribution to the field of data science. It has the potential to enhance the quality and reliability of data used in research and decision-making, thereby fostering the development of more responsible and trustworthy AI systems.

\bibliography{citations}

\section*{Author contributions statement}

A.C.R. and J.L.F. conceived the concept. A.C.R. implemented the framework. A.C.R and V.S. conducted the evaluation. A.C.R. wrote the draft manuscript. S.B. contributed to the design of the responsible AI components. J.W.V., K.X., C.B. and J.T. contributed to the content of all sections of the manuscript. V.S. contributed to the legal and privacy sections. All authors reviewed the manuscript. 

\section*{Additional information}

\subsection*{Competing interests statement}

The authors declare no competing interests.

\end{document}